\documentclass[10pt,twocolumn,letterpaper]{article}
\newcommand{\tabincell}[2]{\begin{tabular}{@{}#1@{}}#2\end{tabular}}
\usepackage{iccv}
\usepackage{times}
\usepackage{epsfig}
\usepackage{graphicx}
\usepackage{amsmath}
\usepackage{amssymb}

\usepackage[breaklinks=true,bookmarks=false]{hyperref}

\iccvfinalcopy 

\def\iccvPaperID{****} 

\ificcvfinal\fi

\begin{document}

\title{Multi-modal Multi-label Facial Action Unit Detection with Transformer}

\author{Lingfeng Wang, Shisen Wang, Jin Qi\\
University of Electronic Science and Technology of China\\
Chengdu, China\\
{\tt\small \{wanglingfeng,shisenwang\}@std.uestc.edu.cn}
}

\maketitle
\ificcvfinal\thispagestyle{empty}\fi

\begin{abstract}
   Facial Action Coding System is an important approach of  facial expression analysis.This paper describes our submission to the third Affective Behavior Analysis (ABAW) 2022 competition. We proposed a transfomer based model to detect facial action unit (FAU) in video. To be specific, we firstly trained a multi-modal model to extract both audio and visual feature. After that, we proposed a action units correlation module to learn relationships between each action unit labels and refine action unit detection result. Experimental results on validation dataset shows that our method achieves better performance than baseline model, which verifies that the effectiveness of proposed network.
\end{abstract}

\section{Introduction}

Facial affective behavior recognition plays an important role in human-computer interaction \cite{kollias2021analysing}. It allows computer systems to understand human feelings and behaviors, which makes human computer interaction more applicable. Facial Action Coding System (FACS) is a important approach of face representation. FACS deconstruct facial expressions into individual components of basic muscle movement, called Action Units (AUs).The task of FAU detection can be formulated as a multi-label binary classification problem.

Since each facial action are defined at specific region within a short duration. So spatial and temporal attention could help model to extract facial features. Moreover, We noticed that the AU labels are highly dependant to each another. Exploiting the correlation between labels could help to boost multi-label classification performance.

In the challenge for Affective Behavior Analysis in-the-wild (ABAW) Competition \cite{kollias2022abaw,kollias2021analysing, kollias2020analysing,kollias2021distribution,kollias2021affect,kollias2019expression,kollias2020face,Kollias_2019,8014982}, the organizers collect a large scale in-the-wild database Aff-Wild2 to provide a benchmark for valence-arousal (VA) estimation, expression (Expr) classification, action unit (AU)
detection, multi-task-learning (MTL) tasks respectively. 

In this paper, we describe our approach for Action Unit Detection challenge in the ABAW 2022 competition. Firstly, we designed a visual spatial-temporal transformer based model and a convolution based audio model to extract action unit feature. Secondly, inspired by the relationship between FAUs, we proposed a transformer based correlation module to learn correlation between action unit labels.


\begin{figure*}
\begin{center}
\includegraphics[width=0.9\linewidth]{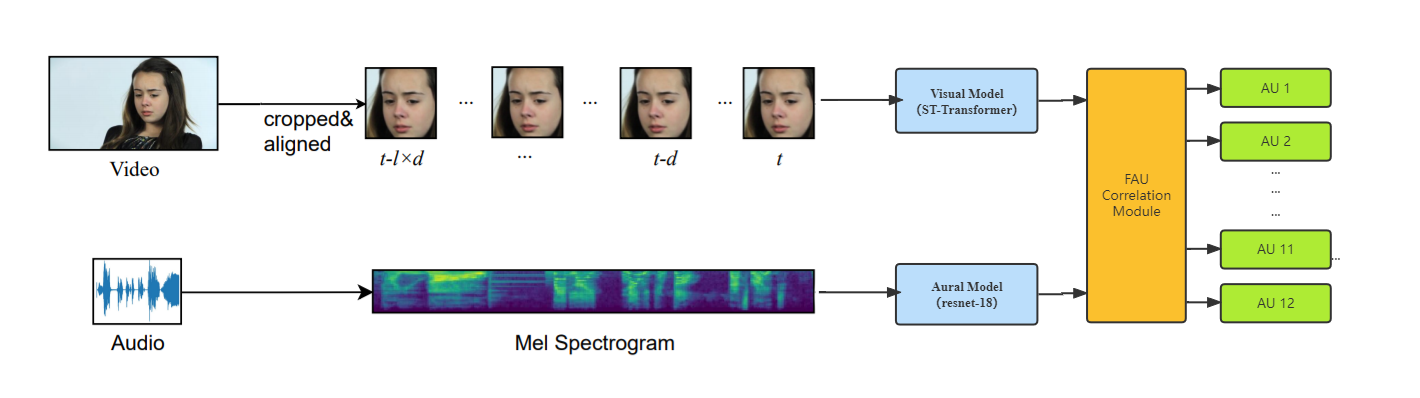}
\end{center}
   \caption{Framework for multi-task affective behavior analysis model.}
\label{fig:model}
\end{figure*}

\begin{figure*}
\begin{center}
\includegraphics[width=0.9\linewidth]{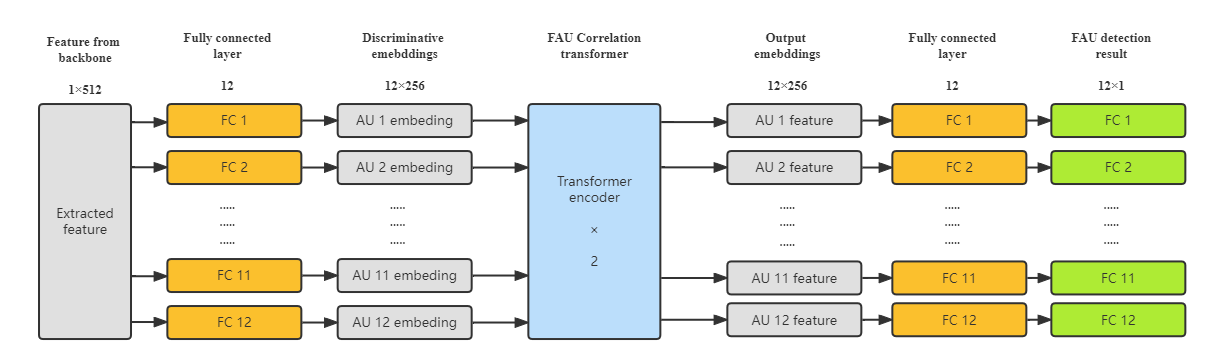}
\end{center}
   \caption{Framework for correlation module.}
\label{fig:mean_teacher}
\end{figure*}

\section{RELATED WORKS}

Previous studies on the Aff-Wild2 have proposed some effective facial action unit detection models. Kuhnke \etal~\cite{Kuhnke_2020} proposed a two-stream aural-visual network to combine vision and audio information for emotion recognition and achieves superior performance. Yue Jin \etal~\cite{MTMSN} proposed a transformer based model to merge feature from visual and audio model in sequence. Since transformer proposed by Vaswani \etal~\cite{vaswani2017attention} has achieved the state of the art performance in many tasks, more and more researchers exploit transfomer for affective behavior analysis. zhao \etal~\cite{zhao2021former} proposed a transformer with spatial and temporal attention for facial expression analysis. Jacob \etal~\cite{jacob2021facial} proposed a transformer correlation module to learn relationship between action units.
Inspired by their work, in this paper we proposed a transformer based aural-visual model to detect action units.

\section{METHODOLOGY}
\subsection{ Framework}
Figure~\ref{fig:model} shows the framework of our multi-task affective behavior analysis model. All the video in the competition dataset are splitting into image clips and audio streams. These streams are pre-processed and then synchronously fed into the aural-visual model. 
For the Visual stream, the input frames are cropped facial region images. These facial crops are all aligned according to 5 point template (eye centers, nose tip, outer mouth corners).Each input clip contains l frames and the frames are sampled with dilation d. Here we choose clip length l = 16 and dilation d = 3. As for audio stream, we compute a mel spectrogramm for all audio stream extracted from the video using TorchAudio package. For each clip, spectrogram is cut into a smaller sub-spectrogram with the center of sub-spectrogram aligning with the current frame at time t.

The two stream are pass through Aural and Visual model respectively. We employ spatial-temporal transformer\cite{zhao2021former} to extract spatio-temporal information from visual stream as well as Resnet-18 \cite{he2016deep}  for mel spectrogram analysis. Finally, the outputs of both sub-models are merged in AU Correlation module and give the joint prediction of action units.

\subsection{Spatial-Temporal transformer}
We use ST transformer from \cite{zhao2021former} as visual backbone. It mainly consists of a convolutional spatial transformer (CS-Former) and a temporal transformer (T-Former).The CS-Former takes still frame as  input and extract spatial facial features, and then pass through T-Former in sequence and generate final feature representation. The convolutional spatial transformer (CS-Former) consists of five resnet \cite{he2016deep} basic blocks and two spatial transformer encoders, while the temporal transformer (T-Former) consists of 3 temporal encoders. The transformer encoder has two main
components: MultiHead Attention and Feed Forward Networks\cite{vaswani2017attention}.
 
\subsection{AU Correlation module}
 Figure~\ref{fig:model} shows the framework of proposed AU Correlation module. We use 12 independent fully connected layer branch to extract features for 12 AU labels. The discriminative features from the of 12 AU branches are provided as input to the AU correlation module. Here we use transformer encoders as as correlation module and the discriminative features as input embeddings for transformer. Relationships between FAU features are learnt in the transformer encoder. The output of transformer encoder are passed to 12 fully connected layer to predict the labels. 

\subsection{Loss Function}

Facial AU detection can be regarded as a multi-label binary classification problem. However, action unit samples in Aff-Wild2 dataset suffer from class imbalance problem. We use binary cross entropy loss (BCE) loss with position weight to tackle the challenge. The position weights here is proportional to the ratio of positives in the total number for each AU class in training set. Weighted BCE allows model to achieve trade-off between recall and precision.

\begin{equation}
L_{BCE} = \mathbb{E}[-\sum (w_{i}t_{i} \cdot log p_{i} + (1-t_{i})\cdot log(1-p_{i}))] \label{XX}
\end{equation}

\section{ EXPERIMENTAL}
\subsection{Dataset}
Proposed model is trained on the large-scale in-the-wild Aff-Wild2 dataset only. This dataset contains 564 videos with frame-level annotations for valence-arousal estimation, facial action unit detection, and expression classification tasks. As for action unit detection task, Aff-Wild2 dataset provide 305 training and 105 validation samples. We use the official provided cropped and aligned images in the Aff-wild2 dataset directly. 

\subsection{Training Setup}
Model is trained with official train split dataset only. As for visual branch, we trained spatial transformer model firstly. After that, we freeze the parameters of the spatial transformer and train the temporal transformer. At the same time, audio model is trained independently. Finally, we combine visual branch and audio branch and train joint model. Models are optimized using Adam optimizer and a learning rate of 0.0005. Random brightness augmentation is applied for each input clip. The mini-batch size is set to 64. 

\subsection{Result}
In order to analyze the effects of our proposed framework design, we conduct ablation studies to compare performance with or without proposed components. The results can be seen in Table I. The usage of correlation module allows the model to learn multi-label relationships and refine classification result. For aural branch and Visual model, the use of correlation module can boost performance by 0.21 and 0.22, respectively. And the use of the temporal transformer can improve the F1 score by 0.21. Moreover, if the aural branch and visual branch are both employed, the F1 score can reach 0.518.
\begin{table}
	\begin{center}
		\begin{tabular}{|l|c|c|c|}
			\hline
			Method & \tabincell{c}{AU (F1)}\\
			\hline\hline
			Competition Baseline & 0.39 \\
			Audio(wo CM) & 0.323\\
			Audio   & 0.344\\
			Visual spatial model(wo CM) &  0.455 \\
			Visual spatial model &  0.479 \\
			Visual spatial temporal model &  0.501 \\
			Joint Aural and Visual &	 0.518\\
			\hline
		\end{tabular}
	\end{center}
	\caption{performance of models on official validation set, CM is short for correlation module}
\end{table}

\section{CONCLUSION}
This paper describe an effective facial action unit detection model by developing a transformer architecture. Our key idea is to firstly develop a multi-modal model to detect action unit. Then we employ correlation module to learn relationship between Action units for further improving the recognition performance. Experimental results on validation dataset show that our model outperforms competition baseline, which verifies the effectiveness of proposed method.

{\small
	\bibliographystyle{ieee_fullname}
	\bibliography{egbib}
}
\end{document}